\def\paperID{16453}
\def\confName{CVPR}
\def\confYear{2024}

\def\paperTitle{Neural Visibility Field for Uncertainty-Driven Active Mapping}

\def\authorBlock{
    Shangjie Xue \quad
    Jesse Dill \quad
    Pranay Mathur \quad
    Frank Dellaert \quad
    Panagiotis Tsiotras \quad
    Danfei Xu\\
    Georgia Institute of Technology \\
    {\tt\small \{xsj, jdill33, pranay.mathur, frank.dellaert, tsiotras, danfei\}@gatech.edu}
}

\newif\ifreview 
\newif\ifarxiv \newcommand{\arxiv}{\arxivtrue}
\newif\ifcamera 
\newif\ifrebuttal 

\arxiv

\pdfoutput=1
\documentclass[10pt,twocolumn,letterpaper]{article}
\ifreview \usepackage[review]{cvpr} \fi
\ifarxiv \usepackage[pagenumbers]{cvpr} \fi
\ifrebuttal \usepackage[rebuttal]{cvpr} \fi
\ifcamera \usepackage{cvpr} \fi

\usepackage{graphicx}
\usepackage{algorithm}
\usepackage{amsmath}	
\usepackage{amssymb}	
\usepackage{booktabs}
\usepackage{times}
\usepackage{microtype}
\usepackage{epsfig}
\usepackage[table,xcdraw,dvipsnames]{xcolor}
\usepackage{caption}
\usepackage{float}
\usepackage{placeins}
\usepackage{color, colortbl}
\usepackage{stfloats}
\usepackage{enumitem}
\usepackage{tabularx}
\usepackage{xstring}
\usepackage{multirow}
\usepackage{xspace}
\usepackage{url}
\usepackage{subcaption}
\usepackage{xcolor}
\usepackage[hang,flushmargin]{footmisc}
\usepackage{wrapfig}
\usepackage{lipsum}
\usepackage[accsupp]{axessibility}

\usepackage{graphicx} %
\usepackage{bm}
\usepackage{amsmath,amssymb,amsfonts}
\usepackage[noend]{algpseudocode}
\ifcamera \usepackage[accsupp]{axessibility} \fi

\usepackage{amsmath}
\DeclareMathOperator*{\argmax}{arg\,max}

\ifarxiv  \fi

\newcommand{\R}[1]{{%
    \textbf{%
        \ifstrequal{#1}{1}{\textcolor{red}{R#1}}{%
        \ifstrequal{#1}{2}{\textcolor{blue}{R#1}}{%
        \ifstrequal{#1}{3}{\textcolor{magenta}{R#1}}{%
        \ifstrequal{#1}{4}{\textcolor{teal}{R#1}}{%
                           \textcolor{cyan}{R#1}%
        }}}}%
    }%
}}

\usepackage{xr-hyper}

\makeatletter
\newcommand*{\addFileDependency}[1]{
  \typeout{(#1)}
  \@addtofilelist{#1}
  \IfFileExists{#1}{}{\typeout{No file #1.}}
}

\makeatother

\definecolor{cvprblue}{rgb}{0.21,0.49,0.74}
\usepackage[pagebackref,breaklinks,colorlinks,citecolor=cvprblue]{hyperref}
\usepackage[capitalize]{cleveref}
\crefname{section}{Sec.}{Secs.}
\crefname{table}{Table}{Tables}
\crefname{figure}{Fig.}{Figs.}

\usepackage{wrapfig}

\renewcommand{\d}{\mathrm{d}}

\newcommand{\revise}[1]{{#1}}

\newcommand{\skipsupp}[1]{{#1}}

\begin{document}
\title{\paperTitle}
\author{\authorBlock}
\maketitle

\begin{abstract}

This paper presents Neural Visibility Field (NVF), a novel uncertainty quantification method for Neural Radiance Fields (NeRF) applied to active mapping. 
Our key insight is that regions not visible in the training views lead to inherently unreliable color predictions by NeRF at this region, resulting in increased uncertainty in the synthesized views. 
To address this, we propose to use Bayesian Networks to composite position-based field uncertainty into ray-based uncertainty in camera observations. Consequently, NVF naturally assigns higher uncertainty to unobserved regions, aiding robots to select the most informative next viewpoints. Extensive evaluations show that NVF excels not only in uncertainty quantification but also in scene reconstruction for active mapping, outperforming existing methods. More details can be found at \url{https://sites.google.com/view/nvf-cvpr24/}.

\end{abstract}

\section{Introduction}
\label{sec:intro}

\begin{figure}
    \centering
    \includegraphics[width=\linewidth]{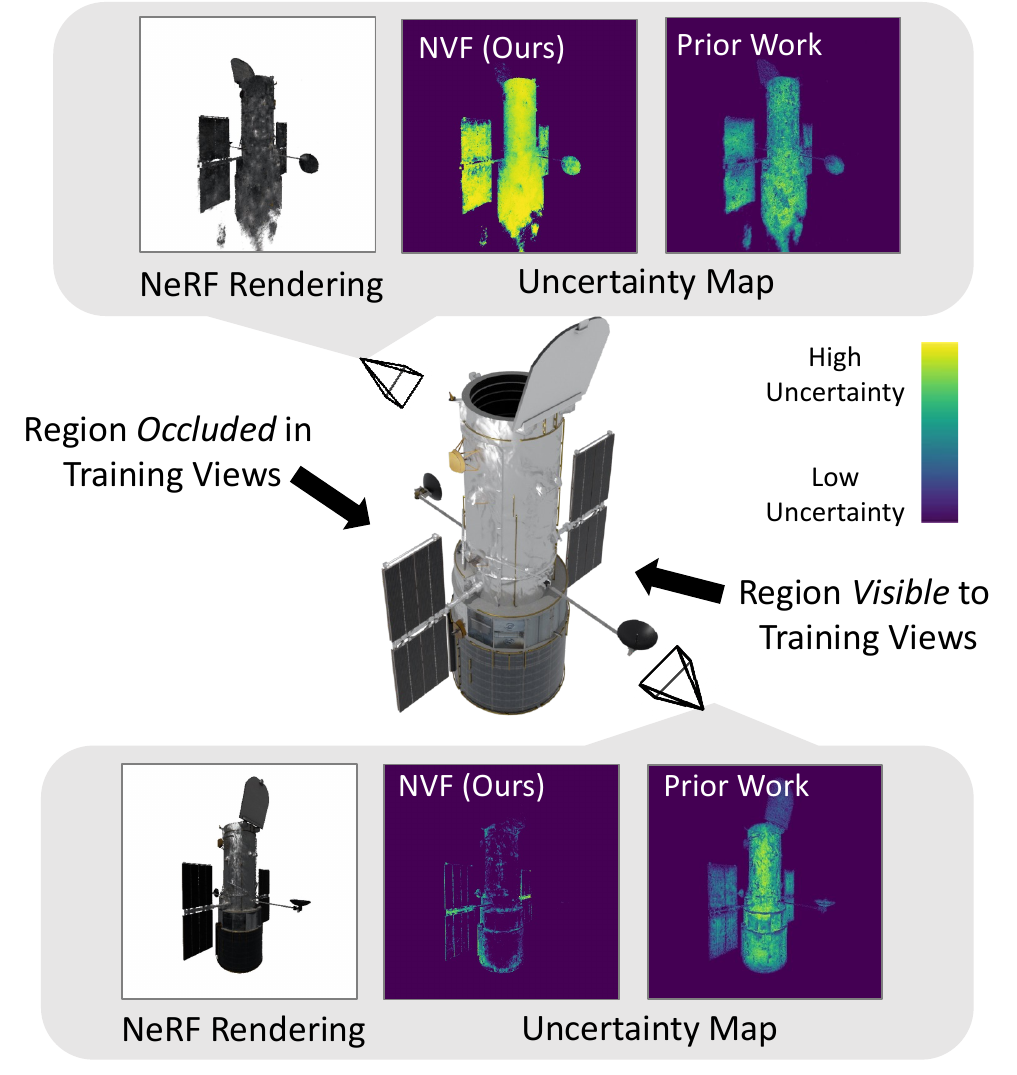}
    \caption{Neural Visibility Field (NVF) is an uncertainty estimation framework for NeRF that accounts for \emph{visibility}: whether a region is covered by the training views of a NeRF. Visible regions should have low uncertainty (bottom row), and unobserved should have high uncertainty (top row). In this paper, we show that many existing methods in NeRF uncertainty quantification can be viewed as special cases of our framework, and NVF outperforms them empirically in uncertainty quantification and active mapping tasks.}
    \label{fig:teaser}
    \vspace{-13pt}
\end{figure}

Active 3D reconstruction plays a pivotal role in robotics. The challenge lies in enabling the robot to precisely reconstruct a target using the fewest views possible. 
Consider the example, illustrated in Figure~\ref{fig:teaser}, where the agent's objective is to thoroughly explore an unknown object (the Hubble telescope).
To achieve this, the robot assesses the uncertainty of potential views, choosing actions that significantly diminish this uncertainty. 
A crucial aspect of this process is the representation of the scene. It should not only facilitate high-quality reconstruction but also be cognizant of uncertainties.

Recently, implicit scene representations, notably NeRF~\cite{mildenhall2021nerf} have shown remarkable ability in high-quality scene reconstructions. The result has motivated applying NeRF for active reconstruction~\cite{pan2022activenerf,yan2023active,zhan2022activermap}. However, due to the opaque nature of neural networks, estimating the uncertainty of NeRF remains challenging. Previous works have developed various proxy measurements to represent the uncertainty in NeRF, in which they aim to maximize the NeRF’s reconstruction accuracy and geometric faithfulness to the scene. However, these approaches neglect a crucial factor to optimize for, namely, visual coverage.

In active reconstruction, an agent makes a tradeoff between exploring new areas of a scene and revisiting previously explored ones. %
Since NeRF is a multiview reconstruction method, a natural strategy is to explore regions that have not been observed by previous views and have these regions hold a high degree of uncertainty. %
Surprisingly, prior methods have largely failed to account for visibility and instead focus on estimating uncertainty via density or NeRF-predicted position-based RGB variance.
Another gap in prior research is the integration of position-based uncertainty factors (e.g., emitted color, opacity, and visibility) into ray-based observation uncertainty. Previous approaches typically employ a simple (weighted) average or sum of position-based uncertainties to approximate the observation uncertainty. However, these methods often lack a solid theoretical foundation and can underperform in complex scenarios. 

To address these challenges, we propose Neural Visibility Field (NVF). Our key insight is that if a region has never been visible in the training views, the color prediction for this point by NeRF is unreliable. To effectively integrate this location-based uncertainty into ray-based camera observations, we view NeRF through the lens of a Bayesian Network. Within this framework, the distribution of a color along a ray can be interpreted as a Gaussian mixture Model.
Subsequently, we calculate the entropy of the GMM and employ it as a cost function, guiding the agent to select the next best view for active mapping. 
We observed that all previous methods can be interpreted as specific approximations within our proposed theoretical framework, yet, they consistently overlook a crucial aspect, namely, visibility.

Our evaluation of the proposed approach is multi-faceted and spans a range of environments, encompassing objects, indoor rooms, and spaces. 
We illustrate how our method offers a superior metric for assessing uncertainty in NeRF. We also apply our approach to active mapping tasks. Specifically, we demonstrate that employing our metric in Next-Best-View (NBV) planning facilitates the planning of trajectories that not only enhance reconstruction quality but also maximize visual coverage of the scene. 
As a result, our proposed method demonstrates significant improvements over these prior approaches in experimental evaluations. %
To summarize, our main contributions are:
\begin{itemize}
    \item 
    We propose a principled uncertainty estimation method for NeRF that takes into account visibility,
    called Neural Visibility Field (NVF).
    \item 
    We provide a unified lens of prior methods in uncertainty estimation for NeRF using NVF.
    \item
    We apply the NVF framework to active mapping tasks demonstrating superior performance compared to the existing state-of-the-art methods.

\end{itemize}

\section{Related Work}

\textbf{Active Mapping.} 
Research on active mapping or NBV selection is a long-studied problem~\cite{connolly1985NBV,Pito1999AST} with the goal of searching for observation poses to create an optimal reconstruction of an environment. Scott et al~\cite{scott2003viewplan} categorizes these approaches as model-based approaches, which utilize knowledge of the geometry and appearance of a scene~\cite{schmid2012view,engelmann2016modelbased}, and model-free approaches, which use information extracted from data gathered online~\cite{connolly1985NBV,Pito1999AST,chen2011survey}.
More relevant are viewpoint selection strategies, including 
frontier-based~\cite{dornhege2011frontier1,Dai2020FastFI}, sampling-based~\cite{tung2018salient,rama2020geometricig,georgios2022geom_uncertainty}, and uncertainty based~\cite{Shen2021StochasticNR,smith2022uncertainty}. 
In particular, our method is inspired by the line of work that uses probabilistic volumetric occupancy to facilitate visibility operations~\cite{kriegal2015volnextbest,isler2016igavr}, which employs the concept of entropy to estimate uncertainty. 

\textbf{Implicit Scene Representation.} 
 Implicit neural fields~\cite{xie2022neuralfields,park2019deepsdf,lars2019occupancy} represent 3D scenes as a continuous differentiable signal parameterized via a neural network. %
 The seminal work of Neural Radiance Fields (NeRFs)~\cite{mildenhall2020nerf} learns a density and a radiance field supervised by multiview 2D images. 
 New views can be queried from a trained NeRF through volumetric rendering. %
 Along this direction, significant progress has been made in %
 novel view rendering~\cite{MartinBrualla2020NeRFIT,mildenhall2020nerf, Uy2023revisit_nerf}, 3D reconstruction~\cite{Azinovic2022CVPR,li2022bnv}, 3D generation~\cite{jain2021dreamfield,poole2022dreamfusion} and videos~\cite{li2021neural,xian2021spacetime, du2021neural, peng2023hbmodel, Li2021Neural3V}. Despite their success, the quality of representation hinges on using a large number of well-posed images which limits their applicability in real-time applications. To counter these problems, recent work has focused on few-shot neural rendering~\cite{yang2023freenerf, chibane2021stereo, niemeyer2021regnerf,deng2021depth}, handling unknown or noisy camera pose estimates~\cite{wang2022nerf,lin2021barf}, using heuristic camera placement strategy \cite{kopanas2023improving}, or adding a notion of uncertainty~\cite{lee2022uncertainty,smith2022uncertainty, Brualla2021nitw, jin2023neu} to quantify information gain for next-best-view selection.

\textbf{Uncertainty Estimation for NeRF.} %
This work focuses on quantifying the epistemic uncertainty of a NeRF model to determine the next best view for improving its reconstruction. Direct approaches such as ensemble-based methods~\cite{sunderhauf2023density} are conceptually simple but computationally expensive or require prior data collection~\cite{hoffman2023probnerf, jin2023neu}. %
Our method improves upon and unifies a recent line of work~\cite{pan2022activenerf,neurar,lee2022uncertainty,zhan2022activermap,yan2023active,smith2022uncertainty}. 
ActiveNerf~\cite{pan2022activenerf} and NeurAR~\cite{neurar} model RGB color distribution at a specific spatial point as a Gaussian distribution, and directly use NeRF to predict its variance. However, the predicted variance tends to be inaccurate in instances where a region has never been visible from the training views.
In comparison,~\cite{lee2022uncertainty, zhan2022activermap, yan2023active, smith2022uncertainty} ignore the spatial RGB uncertainty, and approximating the entropy through the probability of occupancy by using NeRF's density prediction. In particular, \cite{lee2022uncertainty} treats the sampled points in volumetric rendering that are displayed by pixels as discrete random variables and computes the entropy based on it.
In \cite{zhan2022activermap,yan2023active}, the entropy is approximated by utilizing the probability of a ray being occluded at a point.
However, it is worth noting that a remaining gap exists in all previous methods as they lack theoretical grounding for bridging the position-based uncertainty or occupancy uncertainty with ray-based observation uncertainty. In our work, we proposed a theoretically principled method based on Bayesian Network to address this challenge. 
Moreover, a crucial aspect of uncertainty estimation is that if a region is never visible by any of the previous views, the NeRF prediction at this region is not reliable, and high uncertainty should be associated with these regions, yet this aspect is overlooked by all relevant previous works. Our proposed theoretical framework enables us to properly model this aspect through visibility, and all previous work could be viewed as special cases under our proposed framework while lacking certain key aspects.

\label{sec:related}

\section{Method}
\label{sec:method}

Active mapping aims to reduce uncertainties in the reconstructed map and achieve visual coverage of the entire scene. This is achieved by assessing the current uncertainties in the reconstructed map and predicting the potential information gained from proposed viewpoints. 
However, the challenge arises when utilizing NeRF for active mapping.
The opaque and complex nature of neural networks presents significant challenges for accurately quantifying uncertainty within NeRF. 
Although several methods have been proposed to approximate uncertainty in NeRF, they often lack a theoretical foundation and may underperform in complex scenarios. Our key insight is that if a region has never been visible in the training views, the color prediction for this point by NeRF is inherently unreliable. 
To effectively incorporate this position-based uncertainty into camera observations, we propose to use a Bayesian network. This allows for the seamless integration of uncertainty from the implicit field into the observed image’s uncertainty. %
In this section, we will start with a review of NeRF, followed by a detailed explanation of how to model NeRF's volume rendering process using a Bayesian network. Subsequently, we will delve into the integration of visibility aspects within the framework. Finally, we will discuss the application of this framework to active mapping.

\subsection{Problem Formulation}

A NeRF \cite{mildenhall2021nerf} is defined as an implicit function \( F_{\Theta}:(\bm{x}, \bm{d})\rightarrow (\bm{c}, \sigma) \), where \( \bm{x} \) represents the 3D position, \( \bm{d}=(\theta, \phi) \) the viewing direction, \( \bm{c} \) the emitted RGB color at \( \bm{x} \), and \( \sigma \) the volume density at \( \bm{x} \). The volume density function \( \sigma(\bm{x}) \) is a differentiable measure of the probability that a ray is occluded at position \( \bm{x} \). Considering a ray \( \bm{r}(t) = \bm{o} + t \bm{d} \) with near and far bounds \( t_n, t_f \), the observed color at the ray's origin is given by:
\begin{gather}
C(\bm{r}) = \exp\left(-\int_{t_n}^{t_f} T (t)\sigma(\bm{r}(t))\bm{c}(\bm{r}(t), \bm{d}) \d t\right) \label{eq:orig_vr2} \\
T(t) = \exp\left(-\int_{t_n}^{t} \sigma(\bm{r}(s)) \d s\right) \label{eq:trans}
\end{gather}
where $T(t)$ is the transmission probability from \( t_n \) to \( t \) without occlusion. 
Empirically, the algorithm approximates the integral with $N$ discrete samples $\{t_i\}_{i=0}^{N-1}$ along the ray, where \( t_0 \) denotes the ray's origin where the camera is located. Consequently, Eq.~\eqref{eq:orig_vr2} becomes:
\begin{equation}
\hat{C}(\bm{r}) = \sum_i w_i \bm{c}(t_i) \text{,  where } w_i = \alpha_i\prod_{j=0}^{i-1} (1-\alpha_i), \label{eq:orig_vr}
\end{equation}
where \(s_i = t_{i+1} - t_i\) denotes the distance between two adjacent sampled points along the ray, \(\alpha_i = 1 -\exp(- s_i\sigma(t_i) )\) is the alpha value in alpha composition, which can also be viewed as the probability of occlusion at the \(i\)th point.

\subsection{Volume Rendering as Bayesian Network}

While NeRF synthesizes novel views, NeRF cannot estimate the uncertainty in the views. We introduce a method that composites position-based uncertainty into ray-based uncertainty using a probabilistic graphical model. 
This framework enables the integration of visibility factors into the uncertainty estimation process (see~Sec.~\ref{sec:3.3}). 
We consider the observed color along a ray, \( C(\bm{r}) \), as a random variable instead of a constant. In this subsection, we detail the computation of this variable's distribution by using a Bayesian network to model the volume rendering process. 
We use a binary random variable \( D_i \) to denote whether the ray is occluded in the interval \( [t_i, t_{i+1}] \) ($D_i=1$ for occluded, $D_i=0$ for transparent).
The continuous random variable \( C_i \) then represents the emitted color at \( t_i \) in the direction \( \bm{d} \), and \( \bm{Z}_i \) is a continuous random variable for the observed color at \( t_i \). Here, \( \bm{Z}_0 \) corresponds to the camera's observed color at the origin, and hence, the goal is to compute $p(\bm{z}_0)$. 
Notice that, although both $\bm{C}_i$ and $\bm{Z}_i$ represent colors, their difference lies in the objects they represent: $\bm{C}_i$ represents the color distribution associated with a specific position in $\mathbb{R}^3$, whereas $\bm{Z}_i$ corresponds to the color distribution of a camera observation, associated with a particular ray.
For simplicity, we omit the ray index \(\bm{r}\) for \(D_i\), \(\bm{C}_i\), and \(\bm{Z}_i\) in Sections 3.2 and 3.3 below.

Note that the value of \( \bm{Z}_i \) only depends on \( D_i \), \( \bm{C}_i \), and \( \bm{Z}_{i+1} \). Specifically, if the interval \( [t_i, t_{i+1}] \) occludes the ray, \( \bm{Z}_i \) assumes the emitted color \( \bm{C}_i \); otherwise, \( \bm{Z}_i \) equals \( \bm{Z}_{i+1} \), as the interval is transparent. The conditional probability \( p(\bm{z}_i|D_i, \bm{c}_i, \bm{z}_{i+1}) \) is thus:
\begin{equation}
p(\bm{z}_i|D_i, \bm{c}_i, \bm{z}_{i+1}) = \begin{cases} 
\delta(\bm{z}_i - \bm{c}_i), & \text{if } D_i = 1, \\
\delta(\bm{z}_i -  \bm{z}_{i+1}), & \text{otherwise,}
\end{cases} \label{eq:condition}
\end{equation}
where $\delta(\cdot)$ is the Dirac delta function.
Thus, the volume rendering process could be modeled as a hybrid Bayesian network (illustration included in the Appendix). %
Moreover, we can express the marginal probability of \( \bm{z}_i \) using the following recursion:
\begin{equation}
p(\bm{z}_i) = \alpha_i p(\bm{c}_i) + (1-\alpha_i) p(\bm{z}_{i+1}).
\label{eq:recursive}
\end{equation}
Note that this formulation utilizes the relationship $ P(D_i=0) = 1 - \alpha_i = \exp\left(-\sigma_i s_i\right) $. where \( \alpha_i \) was previously defined as the probability of occlusion at the \(i\)th point along the ray.
If we assume \( p(\bm{c}_i) \) is a Gaussian distribution with mean \( \bm{\mu}_{\bm{c}_i} \) and covariance \( \bm{Q}_{\bm{c}_i} \), as predicted by the NeRF model (see Sec. 3.4 for details), by using recursion Eq.~\eqref{eq:recursive}, the marginal probability of \( \bm{z}_0 \) is computed as a Gaussian mixture model (GMM):
\begin{equation}
p(\bm{z}_0) = \sum_i w_i \mathcal{N}(\bm{\mu}_{\bm{c}_i}, \bm{Q}_{\bm{c}_i}),
\end{equation}
with \( w_i = \alpha_i\prod_{j=1}^{i-1} (1-\alpha_i) \). 
The distribution of the camera's observation, \( p(\bm{z}_0) \), implies that \( E[\bm{z}_0] = \sum_i w_i \bm{\mu}_{\bm{c}_i} \)
which aligns with the original NeRF's volume rendering expression (Eq.~\eqref{eq:orig_vr}). 

So far, we have developed a framework based on a probabilistic graphical model to bridge position-based uncertainty with ray-based observational uncertainty. 
In the following subsection, we will discuss the integration of the visibility factor into this framework.

\begin{figure*}
    \centering
    \includegraphics[width=0.9\linewidth]{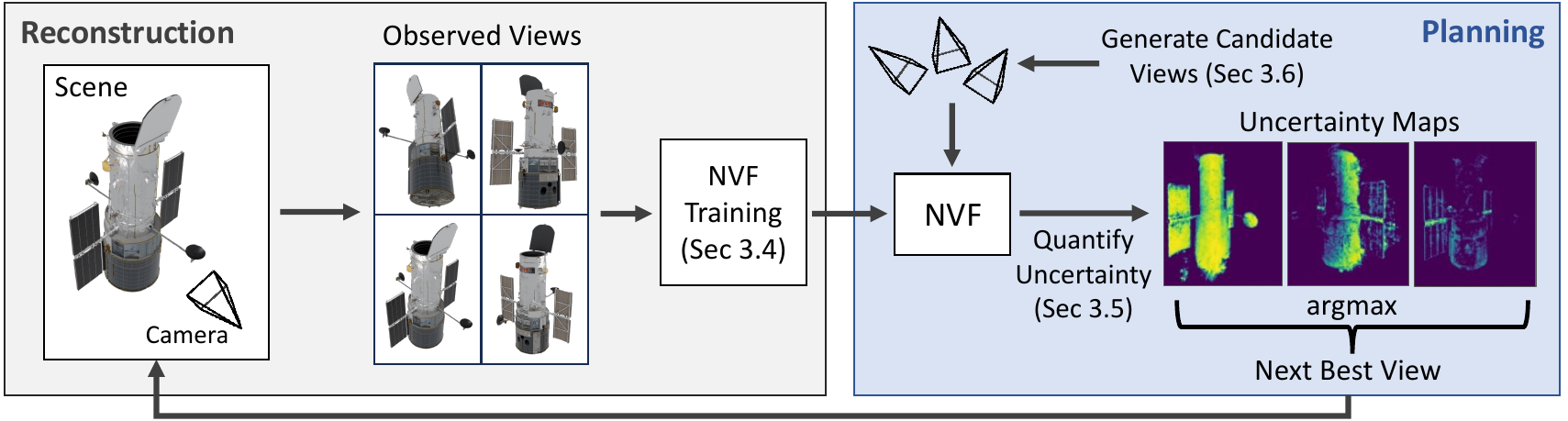}
    \caption{\textbf{Active Mapping with NVF.} Starting with a small set of initial views, a trained NVF is used to quantify uncertainties among sampled candidate views and chooses the view with maximum uncertainty as the next view to be observed by the agent.}
    \label{fig:model}
    \vspace{-5pt}
\end{figure*}

\subsection{Uncertainty with Visibility} \label{sec:3.3}

With the Bayesian network formulation, we can now add visibility into the uncertainty estimation. 
Let a binary random variable $V_i$ represent whether point $i$ is visible to any camera in the training set. When a point is visible (\( V_i = 1 \)), we can rely on NeRF's output for RGB and its variance. 
If a point is unobserved (\( V_i = 0 \)), the NeRF's output at this point becomes unreliable, and we assign a prior color distribution \( \mathcal{N}(\bm{\mu}_0, \bm{Q}_0) \) to it, as follows:

\begin{equation}
p (\bm{c}_i | V_i) = \begin{cases} 
\mathcal{N}(\bm{\mu}_{\bm{c}_i}, \bm{Q}_{\bm{c}_i}), & \text{if } V_i=1, \\
\mathcal{N}(\bm{\mu}_{0}, \bm{Q}_{0}), & \text{otherwise.}
\end{cases} \label{eq:pcv}
\end{equation}
Moreover, for invisible points, density prediction may also lack accuracy. Therefore, we define the conditional probability table for \( P(D_i|V_i) \) as follows
\begin{equation}
\begin{tabular}{c|cc}
\multicolumn{1}{c}{} & \multicolumn{2}{c}{Occlusion $D_i$} \\
Visibility $V_i$ & 1     & 0 \\
\midrule
1     & $\exp(-\sigma_i s_i)$ & $1- \exp(-\sigma_i s_i)$ \\
0     & $\rho_i$ & $1 - \rho_i $ \\ 
\end{tabular} \label{eq:pdv}
\end{equation}
where $\rho_i = (1-\beta) \exp(-\sigma_0 s_i) + \beta \exp(-\sigma_i s_i)$, and 
the hyperparameter \(\beta\) represents the accuracy of occlusion prediction, specifically indicating the likelihood that a prediction about occlusion is correct for points that are invisible. 
In situations where the occlusion prediction is incorrect, we resort to using a constant prior density \(\sigma_0\) to estimate the occlusion probability. This approach helps in adjusting our model's predictions on density, particularly for points not visible to any camera in the training set.

By combining Eqs.~\eqref{eq:trans},~\eqref{eq:pcv},~\eqref{eq:pdv}, the marginal probability of \( \bm{z}_i \) satisfies the recursive formula similar to Eq.~\eqref{eq:recursive}:
$p(\bm{z}_i) = \alpha_i^* \mathcal{N}(\bm{\mu}_{\bm{c}_i}, \bm{Q}_{\bm{c}_i}) + (1-\alpha_i^*) \mathcal{N}(\bm\mu_{0}, \bm{Q}_{0})$,
where $\alpha_i^* = \bigl(v_i +(1-v_i)\beta \bigr) \bigl(1-\exp(-\sigma_i s_i)\bigr) + (1-\beta)(1-v_i)(1-\exp(-\sigma_0 s_i))$, and $v_i = P(V_i=1)$ is the probability of point $i$ being visible to at least one camera in the training set. The marginal probability of $p( \bm{z}_0)$ can be computed similarly as follows
\begin{equation}
p(\bm{z}_0) = \sum_i w_i^*v_i \mathcal{N}(\bm{\mu}_{\bm{c}_i}, \bm{Q}_{\bm{c}_i}) + \mathcal{N}(\bm{\mu}_{0}, \bm{Q}_{0}) \sum_i w_i^*(1-v_i),  \label{eq:gmm}
\end{equation}
resulting in a GMM as well, where $w_i^* = \alpha_i^*\prod_{j=1}^{i-1} (1-\alpha_j^*)$.

\subsection{Neural Visibility Field} \label{sec:3.4}

So far, we have established a framework that bridges position-based uncertainty with ray-based observation uncertainty, while also incorporating visibility factors.
Next, we discuss a method to determine visibility \(v_i\), which is the probability that a point \(\bm{x}_i\) is visible to at least one camera in the training set. Let \(\mathcal{P} = \{\bm{p}_1, \bm{p}_2, \ldots\}\) be the set of camera poses in the training set. If point \(\bm{x}_i\) is within the field of view of a camera  \(\bm{p}\) in \(\mathcal{P}\), the visibility of \(\bm{x}_i\) to  camera \(\bm{p}\) can be expressed as \(v_{\bm{p}}(\bm{x}_i) = T^p(t_i^p)\), where \(\bm{x}_i = \bm{o_p} + t_i^p\bm{d_p}\) is on a ray from camera \(\bm{p}\), and \(T_p(t_i^p)\) denotes the probability of the ray being transmitted from \(\bm{o_p}\) to \(t_i^p\) without occlusion, as defined in Eq.~\eqref{eq:trans}.  
Therefore, the probability that point \(\bm{x}_i\) is visible to at least one camera in the set \(\mathcal{P}\) is given by:
\begin{equation}    \label{eq:vis_gt}
v_\mathcal{P} (\bm{x}) = 1 - \prod_{\bm{p} \in \mathcal{P}} (1 - v_p(\bm{x}_i)).
\end{equation}
However, directly computing Eq.~\eqref{eq:vis_gt} during volume rendering is impractical. For each point along a ray \(\bm{r}\), it would require generating an additional ray \(r_p\) from camera \(\bm{p}\) and sampling points along this ray to determine the point's visibility to camera \(\bm{p}\). 
Doing this for all existing views is computationally expensive. To address this, we propose to amortize the cost by training an implicit model to predict the visibility. %

We introduce Neural Visibility Field (NVF), an augmented NeRF that outputs both color uncertainty and visibility. 
The enhanced model is defined as \( F_\Theta: (\bm{x}, \bm{d}) \rightarrow (\sigma, \bm{\mu_c}, \bm{Q_c}, v) \), where \(v\) represents the visibility with respect to the training views, and
\(\bm{\mu_c}\) and \(\bm{Q_c}\) denote the mean and covariance of the color vector, respectively. 
The parameters 
$\bm{\mu_c}$ and $\sigma$ are trained with Mean Square Error loss as in~\cite{mildenhall2020nerf}. 
To train $\bm{Q_c}$, we employ the Negative Log-Likelihood Loss as follows:
\begin{equation}
\mathcal{L}_{cov} = - \sum_{r \in \mathcal{R}}  \log \Big(\sum_i w_i \mathcal{N}\big(C_g(\bm{r}); \bm{\mu_{c_i}}, \bm{Q_{{c}_i}}\big)\Big),
\end{equation}
where $\mathcal{R}$ denotes the set of rays in each batch, and $C_g(\bm{r})$ represents the ground truth color of ray $\bm{r}$. 
For training, we randomly sample points within the scene. 
The ground truth visibility, derived using Eq.~\eqref{eq:vis_gt}, is then utilized to train the visibility head,
using cross-entropy loss. 
Please refer to Supp. for further details on network architecture and training.

\subsection{Active Mapping with NVF}

In this section, we apply the PDF of ray color, derived from~Sec.~\ref{sec:3.4}, for active mapping purposes. Let \( Z^{mn}_{\bm{p}} \) be the color of the ray corresponding to pixel index \( m,n \) from camera pose \( \bm{p} \). The PDF of \( Z^{mn}_{\bm{p}} \), denoted as \( p(\bm{z}^{mn}_{\bm{p}}) \), can be obtained using the formulation provided in Eq.~\eqref{eq:gmm}. We define \(\bm{Z_p}\) as a random variable in \(\mathbb{R}^{H \times W \times 3}\), representing the collective observation of all pixels in an image with height \(H\) and width \(W\). 

The goal of active mapping is to identify a camera pose, denoted as \(\bm{p}^*\), that maximizes the entropy of the observation \(\bm{Z}\) at that pose. This is formally expressed as:

\begin{equation}  \label{eq:i_gain}
\bm{p}^* = \argmax_{\bm{p}} \mathcal{H} (\bm{Z_p} ). 
\end{equation}
Note that we can deduce Eq.~\eqref{eq:i_gain} from the information gain or mutual information 
\(\mathcal{I}(\bm{Z_p}; \bm{M}) = \mathcal{H}(\bm{Z_p}) - \mathcal{H}(\bm{Z_p| \bm{M}})\), where 
\(\bm{M}\) represents the random variable of the entire map. 
This assumes that \(\mathcal{H}(\bm{Z_p| \bm{M}})\) is constant, specifically, that measurement noise remains constant given a known map.

To compute \(\mathcal{H}(\bm{Z_p})\), we initially assume that the color of each pixel is independent of the others. 
Under this assumption, the entropy of \(\bm{Z_p}\) can be calculated as \(\mathcal{H}(\bm{Z_p}) = \sum_{m,n} \mathcal{H}(\bm{Z_p}^{mn})\). 
However, this assumption of independence may not always hold true. 
For instance, when the camera is in close proximity to an object, the pixels in the image are often strongly correlated, particularly since they are measuring points that are spatially close. 
To account for this spatial correlation, we introduce a correction term:
\begin{equation}
\mathcal{H}(\bm{Z_p}) = \sum_{m,n} \Big( \mathcal{H}(\bm{Z_p}^{mn}) - f_{\rm corr}(\mathcal{H}(\bm{Z_p}^{mn}); d_{\bm{p}}^{mn}) \Big)\label{eq:id}
\end{equation}
Here, \(f_{\rm corr}(\mathcal{H}(\bm{Z_p}^{mn}); d_{\bm{p}}^{mn})\) incorporates spatial correlation based on the expected depth \(d_{\bm{p}}^{mn}\). 
Furthermore, we use the upper bound as proposed in~\cite{huber2008entropy} to closely approximate the entropy of the GMM $\mathcal{H}(\bm{Z_p}^{mn})$, as it is known that there is no analytical solution for the entropy of GMM~\cite{huber2008entropy}.
Further details on \(f_{\rm corr}(\mathcal{H}(\bm{Z_p}^{mn}); d_{\bm{p}}^{mn})\) and entropy computation are included in supp material.

Within our theoretical framework for estimating uncertainty in NeRF and active mapping, all prior works, to our best knowledge, can be viewed as special cases. 
Specifically, if we drop the visibility factor, each prior work can be viewed as a specific approximation of our method. 
For instance,
Lee et al~\cite{lee2022uncertainty} focuses only on the discrete random variable, computing the Shannon entropy with \( -w_i \log w_i \), which can be regarded as a simplified version of ours, albeit excluding the differential entropy term
Similarly,~\cite{neurar,pan2022activenerf} uses the weighted average of position-based color variance to approximate the rays-based observation variance, which lacks theoretical grounding 
and ignores the visibility factor in weight computation. 
In addition, Zhan et al~\cite{zhan2022activermap} approximate the entropy of ray-based observation by directly summing the position-based entropy of occlusion, whereas, similarly, 
Yan et al~\cite{yan2023active} use the weighted average of position-based entropy of occlusion to approximate the ray-based observation entropy.

\subsection{Active Mapping Pipeline}\label{sec:3.6}

Here we briefly describe the active mapping pipeline using NVF (illustrated in Fig.~\ref{fig:model}). Please refer to Supp. material for more details. The process starts with training the NVF on a small batch of initial views. We employ two strategies for next view selection. 
In the sampling-based strategy, we sample $N$ views from a prior distribution, estimate their uncertainty using Eq.~\eqref{eq:i_gain}, and select the view with maximum uncertainty. The gradient-based strategy is implemented by adjusting the selected pose through gradient-based optimization, aimed at maximizing entropy, leveraging the inherent differentiability of our uncertainty estimation method. 
Lastly, the agent proceeds to collect and integrate the new observations into the training views to re-train the NVF model and plan the next view.

\section{Experiments}
\label{sec:exp}

\begin{figure*}
    \centering
    \includegraphics[width=0.85\textwidth]{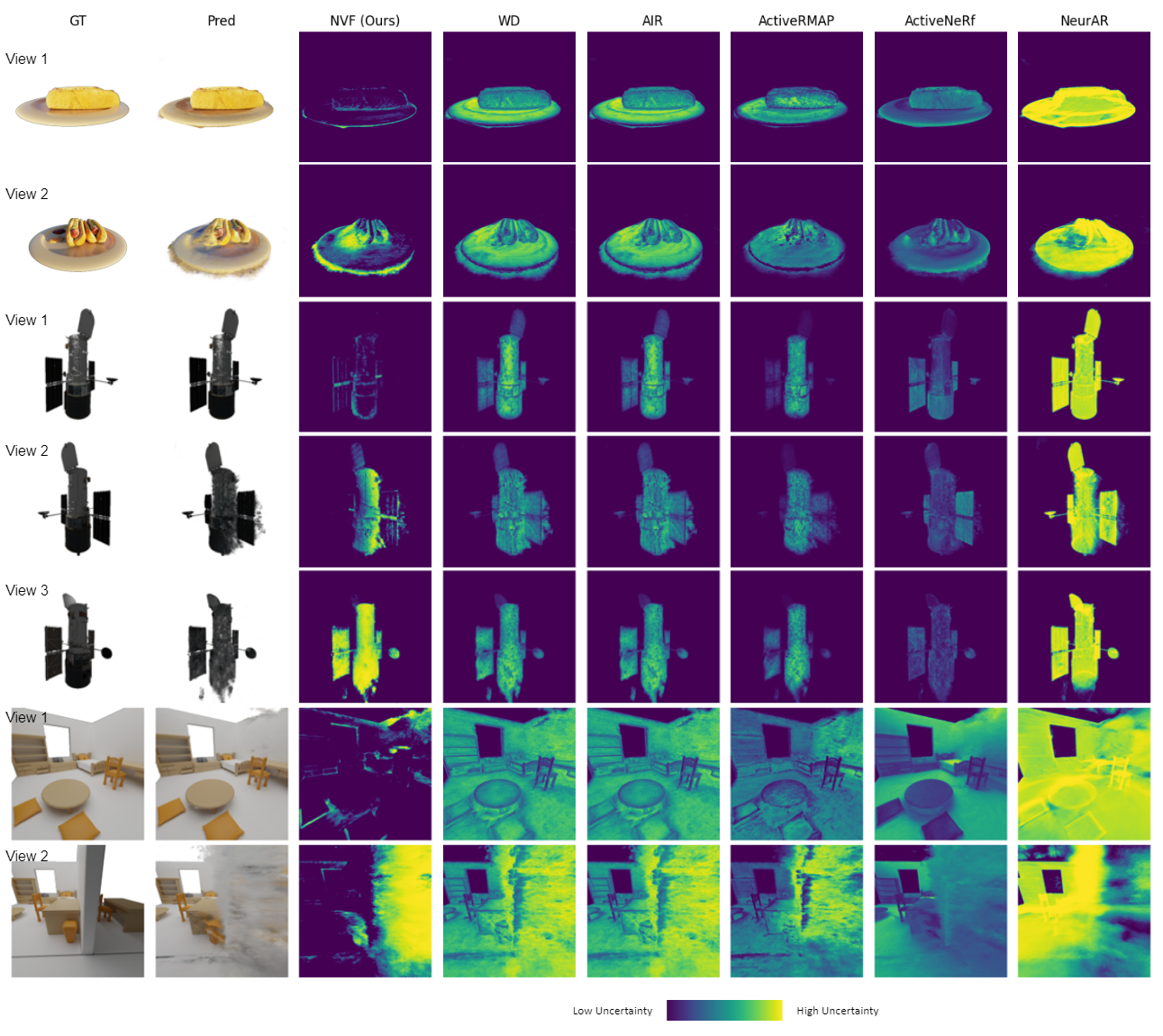}
    \vspace{-10pt}
    \caption{\textbf{Qualitative results of entropy estimation:} NVF assigns a higher entropy to previously unobserved regions while the baselines do not distinguish between the observed (View 1) and unobserved regions (View 2/3). Schematic illustrations of the poses of View 1, 2, and 3 can be found in supp. material. Note that within each method and scene, all rendered views share the same color bar. }
    \label{fig:entropy_qualitative}
    \vspace{-10pt}
\end{figure*}

In this section, we seek to verify our hypothesis that (a) NVF outperforms existing methods in both uncertainty quantification and active mapping both quantitatively and qualitatively; and (b) the visibility term plays a vital role in this result. %

\noindent\textbf{Simulation Environments and Learning Setup.} 
We conduct experiments on three datasets of varying difficulty levels for active mapping: \revise{all assets} from the original NeRF dataset~\cite{mildenhall2021nerf}, the Hubble Space Telescope, and a custom synthetic indoor Room scene. In particular, the Room scene consists of two spaces divided by a wall. Successfully mapping the scene requires traversing both spaces. 
We assume access to a coarse bounding box that contains the region of interest.
All ground truth images used for training NeRF and assessing reconstruction quality were rendered using Blender at a resolution of \(512 \times 512\).
We utilized Instant-NGP~\cite{muller2022instant} as an efficient backbone for all uncertainty estimation methods. 
All NeRF models were trained for 5,000 iterations. For NVF, it first trains the Instant-NGP backbone, freezes its weights, and then trains variance and visibility heads, to ensure the performance improvements are attributed to better entropy estimation instead of a change in the loss function.

\noindent\textbf{Baselines.} 
We compared our method with state-of-the-art NeRF uncertainty quantification and active mapping methods. This includes the weight distribution-based entropy approximation (WD)~\cite{lee2022uncertainty}; occlusion-based entropy approximation (ActiveRMAP)~\cite{zhan2022activermap}; weighted occlusion-based entropy approximation - ActiveImplicitRecon (AIR)~\cite{yan2023active}, and spatial RGB variance-based uncertainty estimation ActiveNeRF~\cite{pan2022activenerf} and NeurAR~\cite{neurar}. 
As discussed earlier, all of these works can be viewed as a special case of our method, 
while missing key aspects that NVF introduces.
In addition, we include an agent that randomly selects views from the candidate poses (Random).

\subsection{Uncertainty Estimation}

\textbf{Setup}. 
We qualitatively compare the uncertainty (entropy) maps produced by our method and the baselines given a set of training views.
For each scene, we design scenarios where only certain regions are visible in the training views. We then train all methods on the same training views and query for uncertainty estimation at an unseen test view. 
An effective uncertainty estimation method should be able to differentiate regions unobserved in the training set, as reconstruction in these areas is noisy and inaccurate. 
For the "Hotdog" scene from the original NeRF dataset, we randomly sample 20 training views from a 90-degree sector above the plate. 
In the Hubble scene, we sample 20 training views from a 90-degree sector on one side, keeping the opposite side of the Hubble out of view. 
For the Room scene, we sample 30 training views oriented toward the back wall, the common wall, and the floor of one of the rooms, ensuring that the other room is unobserved.

\begin{figure*}
    \centering
    \includegraphics[width=0.75\textwidth]{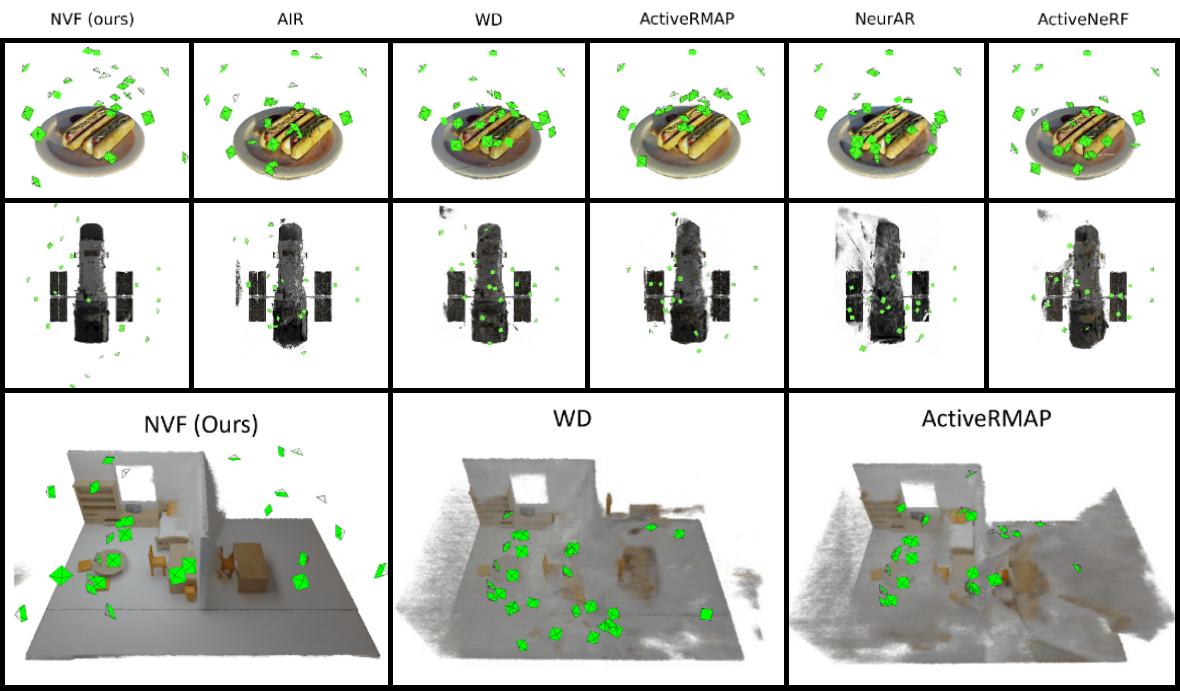}
    \caption{\textbf{Reconstruction results and camera view distribution:} NVF demonstrates superior reconstruction and scene coverage across all datasets in comparison to baselines. 
    \revise{For room scene, only comparable baselines are presented\skipsupp{, full results are provided in supp. material}}.
    }
    \label{fig:mapping_result_fig}
\vspace{-15pt}
\end{figure*}

\noindent\textbf{Results.} 
In the Hubble and Hotdog scenes, as illustrated in Fig.~\ref{fig:entropy_qualitative}, all baseline methods fail to accurately capture the uncertainty in the unobserved areas of the scenes. 
Several baselines assign greater or similar uncertainty to the observed views as compared to the unobserved ones.
A notable example is the Room scene. 
The first view focuses on the room observed in the training view, while the second view targets the common wall between the two rooms. 
The result indicates that our method differentiates between the uncertainties in regions seen in training views and unseen regions by modeling their visibilities.
In Hubble and Hotdog, ActiveNeRF and NeurAR estimate a similar level of uncertainty for both unobserved and observed regions. 
However, in the Room scene, the uncertainty in the unobserved region is estimated to be lower than the observed. 
This shows that in complex scenarios, the uncertainty formulations of ActiveNeRF and NeurAR are ineffective, and such formulation alone is insufficient as guidance to explore unobserved regions.

\subsection{Active Mapping}
\noindent\textbf{Setup.} We deploy active mapping agents with the pipeline described in Sec.~\ref{sec:3.6} with different uncertainty estimation methods. %
To ensure a fair comparison, all methods are evaluated under the same conditions during the comparison, to ensure that the planning is driven solely by the uncertainty estimation. Specifically, all candidate views are uniformly sampled within the space, without any prior constraints (such as the hemisphere constraint employed in ActiveNeRF). This approach ensures that a more accurate uncertainty estimation method will enable the robot to achieve more precise mapping results.
For all original NeRF assets and Hubble scenes, we utilize 3-5 initial views covering only a portion of the scene, to realistically simulate active mapping scenarios. 
The Room scene presents the greatest challenge, with nine initial views sampled from one room, leaving the second room entirely unexplored. All agents start without knowledge of the second room's existence and are expected to discover it through uncertainty estimation and reconstruct the scene in 20 steps. Please refer to Supp. material for more details.

\begin{table*}[htbp]
  \centering
  \caption{Evaluation of Reconstructed Models Using Different Methods for Active Mapping} \label{table:1}
  \vspace{-8pt}
    \begin{tabular}{cccccccccc}
    \toprule
    Scene & Method & PSNR$\uparrow$ & SSIM$\uparrow$ & LPIPS$\downarrow$ & RGB$\downarrow$ & Acc.$\downarrow$ & Comp$\downarrow$ & CR$\uparrow$ & Vis$\uparrow$ \\
    \midrule
    \multirow{7}[2]{*}{\shortstack{NeRF \\ Assets \\ (Avg.)}} & Random & 17.63 & 0.766 & 0.264 & 0.0193 & 0.0426 & 0.0401 & 0.348 & 0.225 \\
        & WD   & 19.91 & 0.807 & 0.227 & 0.0121 & 0.0311 & 0.0204 & 0.479 & 0.466 \\
      & ActiveRMAP & 20.03 & 0.807 & 0.219 & 0.0118 & 0.0292 & 0.0184 & 0.510 & 0.471 \\
      & AIR   & 19.86 & 0.807 & 0.230 & 0.0118 & 0.0290 & 0.0195 & 0.494 & 0.453 \\
      & ActiveNeRF & 18.78 & 0.771 & 0.281 & 0.0157 & 0.0301 & 0.0238 & 0.433 & 0.415 \\
      & NeurAR & 19.58 & 0.755 & 0.286 & 0.0134 & 0.0347 & 0.0251 & 0.452 & 0.424 \\
      & NVF (Ours) & \textbf{23.90} & \textbf{0.890} & \textbf{0.106} & \textbf{0.0045} & \textbf{0.0193} & \textbf{0.0111} & \textbf{0.685} & \textbf{0.532}  \\
    \midrule
    \multirow{7}[2]{*}{Hubble} & Random & 21.76 & 0.778 & 0.265 & 0.0113 & 0.0734 & 0.0262 & 0.329 & 0.291 \\
          & WD    & 24.15 & 0.855 & 0.184 & 0.0039 & 0.0297 & 0.0184 & 0.471 & 0.571 \\
          & ActiveRMAP & 23.34 & 0.835 & 0.205 & 0.0048 & 0.0282 & 0.0162 & 0.465 & 0.570 \\
          & AIR   & 24.63 & 0.862 & 0.182 & 0.0035 & 0.0249 & 0.0140 & 0.525 & 0.586 \\
          & ActiveNeRF & 23.33 & 0.824 & 0.250 & 0.0047 & 0.0355 & 0.0201 & 0.442 & 0.552 \\
          & NeurAR & 25.19 & 0.772 & 0.265 & 0.0030 & 0.0480 & 0.0170 & 0.416 & 0.537 \\
          & NVF (Ours) & \textbf{27.99} & \textbf{0.919} & \textbf{0.100} & \textbf{0.0016} & \textbf{0.0225} & \textbf{0.0110} & \textbf{0.651} & \textbf{0.681} \\
    \midrule
    \multirow{7}[2]{*}{Room} & Random & 12.95 & 0.800 & 0.378 & 0.0563 & 0.1837 & 0.5468 & 0.338 & 0.397 \\
          & WD    & 13.42 & 0.792 & 0.387 & 0.0533 & 0.2893 & 0.5415 & 0.317 & 0.428 \\
          & ActiveRMAP & 13.91 & 0.786 & 0.412 & 0.0411 & 0.2233 & 0.4646 & 0.317 & 0.450 \\
          & AIR   & 15.19 & 0.829 & 0.386 & 0.0307 & 0.2710 & 0.3153 & 0.343 & 0.498 \\
          & ActiveNeRF & 10.69 & 0.733 & 0.434 & 0.0853 & 0.1847 & 0.8181 & 0.292 & 0.338 \\
          & NeurAR & 12.23 & 0.584 & 0.508 & 0.0599 & 0.3948 & 1.2647 & 0.178 & 0.375 \\
          & NVF (Ours) & \textbf{22.83} & \textbf{0.943} & \textbf{0.156} & \textbf{0.0053} & \textbf{0.1132} & \textbf{0.1997} & \textbf{0.464} & \textbf{0.586} \\
    \bottomrule
    \end{tabular}%
    \vspace{-13pt}
\end{table*}%

\noindent\textbf{Evaluation metric.} Our evaluation employs three types of metrics. %
For novel view synthesis quality, evaluations are performed at fixed testing viewpoints. We compare views synthesized by NeRF with ground truth renderings. The errors are quantified using Peak Signal-to-Noise Ratio (PSNR), Perceptual Image Patch Similarity (PIPS), Learned Perceptual Image Patch Similarity (LPIPS)~\cite{zhang2018perceptual}, and RGB loss.  %
For reconstructed mesh quality, we quantitatively evaluate the geometric accuracy of the scene reconstructions. 
We employ the metrics, Accuracy (Acc), Completion (Comp), and Completion Ratio (CR) as proposed in~\cite{Sucar:etal:ICCV2021}. %
\revise{For visual coverage (Vis), we assess the proportion of faces in the ground truth mesh observed without occlusion during the experiments over all faces. The visibility of each face in the mesh is tracked using the ground truth mesh and a rasterizer.}

\begin{table}[t]
\caption{Ablation Studies for Active Mapping with NVF\label{table:2}}
\vspace{-8pt}
\begin{tabular}{ccccc}
\hline
Ablations        & PSNR$\uparrow$ & SSIM$\uparrow$ & LPIPS$\downarrow$ & Vis$\uparrow$ \\ \hline
w/o Vis. & 21.11  & 0.844  & 0.187  & 0.382  \\
w/o Var. & 23.77  & 0.897  & 0.113  & 0.551  \\
Ind. Rays & 20.32  & 0.822  & 0.236  & 0.482  \\
Loose & 22.54  & 0.881  & 0.137  & 0.504  \\
NVF (Ours) & 24.42  & 0.902  & 0.108  & 0.546  \\ \hline
\end{tabular}%
\label{tab:ablation_result_main}%
\vspace{-17pt}
\end{table}

\noindent\textbf{Results.} 
In Tab.~\ref{table:1}, we show the quantitative results of our approach in comparison to other baselines. \revise{For the original NeRF Assets, we only include the average of the results across all scenes due to space limitation\skipsupp{, detailed results are provided in supp. material}}.
Our method significantly outperforms baseline methods achieving higher-quality reconstruction and improved visual coverage. This is especially evident in challenging scenarios such as the Hubble and Room scenes, where our method successfully explores the entire scene and excels across all metrics. In contrast, baseline methods failed to fully explore these scenes, often revisiting previously explored areas (see Fig.~\ref{fig:mapping_result_fig}) due to inadequate uncertainty estimation that overlooks visibility.

\subsection{Ablation studies}
We ablate key components in NVF to examine their role.
First, we negate the visibility factor by presuming all sampled points as visible to the camera, setting the visibility head output to $1$ for any input. Second, we disregard the spatial color variance estimation from NeRF, assuming a constant small uncertainty for all sampled points. Third, we omit the correlation correction factor, treating all rays as independent. Lastly, for entropy computation, we substitute the upper bound proposed by~\cite{huber2008entropy} with a looser bound, treating multiple Gaussians as a single Gaussian following \cite{hershey2007approximating}. \revise{The average results across all scenes are shown in Tab.~\ref{table:2}, highlighting the crucial role in the visibility factor, removing it significantly drops the performance}. We also observe that the correction of independence in Eq.~\eqref{eq:id} ("Ind. Rays." in Tab.~\ref{table:2}), and a tighter upper bound ("Loose") positively impact performance. However, the position-based color uncertainty directly predicted by NeRF ("w/o Var.") plays a less important role, underscoring visibility as the most critical factor in uncertainty estimation for active mapping.

\section{Conclusion}
\label{sec:conclusion}
In this work, we present Neural Visibility Field, a principled approach that accounts for visibility in uncertainty quantification and provide a unifying view of prior research in this direction. We empirically demonstrated that NVF significantly outperforms baselines in reconstruction quality and visual coverage across three scenes with varying levels of complexity.
A limitation of our current active mapping pipeline is that it does not account for the constraints imposed on the planned trajectory of an agent. A possible future direction is to integrate NVF with cost-aware path planning.

\section*{Acknowledgments}
This work is supported by NSF grant 2101250. We thank Mehregan Dor for the feedback on the preliminary version, and the anonymous reviewers for their comments and feedback on our manuscript.

{\small
\bibliographystyle{ieeenat_fullname}
\bibliography{11_references}
}

\ifarxiv \clearpage \appendix \appendix
\section{Method Details}

\subsection{NVF architecture and training details}
\begin{figure*}
    \centering
    \includegraphics[width=0.9\textwidth]{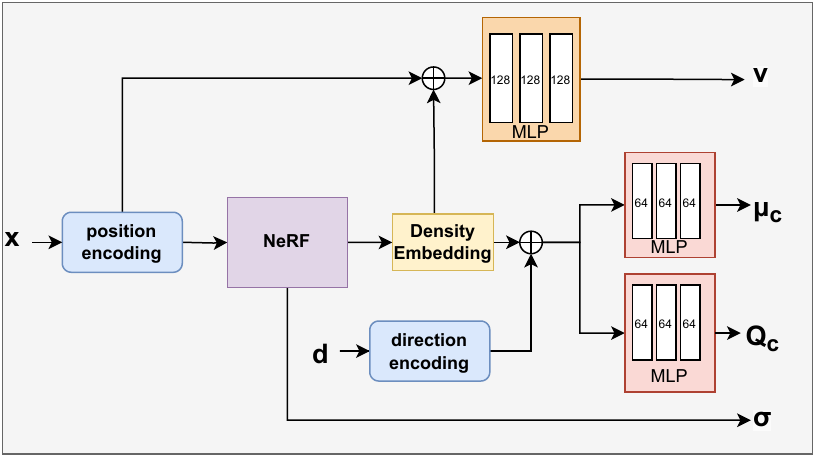}
    \caption{\textbf{NVF Architecture:} The MLP block consists of fully connected layers that use the ReLU activation function. The numbers inside the block denote the size of the layer. The final output from the visibility $(v)$ MLP and RGB $(\mu_c)$ MLP are passed through the sigmoid activation function while the RGB Variance $(Q_c)$ MLP uses softplus activation}
    \label{fig:model_arch}
\end{figure*}
NVF is an augmentation of a NeRF consisting of two additional MLP heads for predicting RGB variance and visibility. Specifically, we implement NVF on top of a nerfstudio~\cite{nerfstudio} implementation of Instant-NGP~\cite{muller2022instant}, where the color MLP head represents $\bm{\mu}_c$. Alongside the color head is a MLP head for RGB variance, outputting a 3x1 vector $\bm{Q}_c$. Similarly, the visibility MLP head is attached alongside the density head. For a visualization of the architecture, see Appendix Fig.~\ref{fig:model_arch}. In practice, we train Instant-NGP, variance, and visibility separately and in sequence. First, we train the NeRF backbone for 5000 iterations using a learning rate of $0.01$ and 4096 rays per batch. Next, the variance head is trained for 500 iterations using a learning rate of $0.001$ and 4096 rays per batch. Finally, the visibility head is trained for 500 iterations using a learning rate of $0.001$ and 65536 samples per batch. %
We train all modules using the Adam optimizer~\cite{kingma2014adam}.

\subsection{Entropy computation details}

\textbf{Joint Entropy of the Camera Observation}. We discuss the details on the computation of the joint entropy $\mathcal{H}(\bm{Z_p})$ as formulated in Eq.~\eqref{eq:id}. For simplicity in this discussion, we denote the joint entropy as $\mathcal{H}(\bm{Z})$ in this section. We model the joint observation of all rays as a Bayesian network, where the observation of each pixel only depends on its adjacent neighboring pixel. Consequently, the joint probability can be factorized as $p(z) = \prod_{mn} p (z_{mn} | z_{m+1,n}, z_{m,n+1})$. Note that for the sake of brevity, boundary terms where a pixel lies at the image edge are omitted here. Then, by applying the chain rule of entropy\cite{cover1999information}, we obtain:

\begin{equation}
\mathcal{H}(\bm{Z}) = \sum_{m,n} \mathcal{H}(\bm{Z}_{mn} | \bm{Z}_{m+1,n}, \bm{Z}_{m,n+1})
\end{equation}
We then apply the inequality $ \mathcal{H}(\bm{Z}_{mn} | \bm{Z}_{m+1,n}, \bm{Z}_{m,n+1}) \leq \mathcal{H}(\bm{Z}_{mn} | \bm{Z}_{m+1,n})$ and $\mathcal{H}(\bm{Z}_{mn} | \bm{Z}_{m+1,n}, \bm{Z}_{m,n+1}) \leq \mathcal{H}(\bm{Z}_{mn} | \bm{Z}_{m,n+1})$. These allow us to derive an upper bound for  $\mathcal{H}(\bm{Z})$:

\begin{equation}
\mathcal{H}(\bm{Z}) \leq \sum_{m,n} \frac{1}{2} \big(\mathcal{H}(\bm{Z}_{m,n} | \bm{Z}_{m+1,n}) + \mathcal{H}(\bm{Z}_{m,n} | \bm{Z}_{m,n+1} ) \big)
\end{equation}

Then we connect the conditional entropy with the correlation between the two adjacent rays.
We let 
\begin{equation}
\rho_{m+1,n} = 1 - \frac{\mathcal{H}(\bm{Z}_{m,n} | \bm{Z}_{m+1,n})}{\mathcal{H}(\bm{Z}_{m+1,n})}
\end{equation}
where $\rho$ is as a measure of correlation. Specifically, a $\rho$ value closer to $1$ indicates $\bm{Z}_{m,n}$ and $\bm{Z}_{m+1,n}$ are strongly correlated, whereas a $\rho$ tends to $0$ suggests they are not correlated. It is important to note that this definition of correlation, based on entropy, differs from the widely recognized Pearson correlation coefficient, and is commonly used in quantum information\cite{vedral2002role}. Consequently, we can obtain the upper bound:

\begin{equation}
\mathcal{H}(\bm{Z}) \leq \sum_{m,n} (1-\rho_{mn}) \mathcal{H}(\bm{Z}_{mn})
\end{equation}

We assume that the correlation between two points in the scene can be modeled as a function of their spatial distance, a concept commonly referred to as the correlation function in statistical physics \cite{sethna2021statistical}. We adopt a truncated least-square form for this correlation function.

\begin{equation}
\rho(x) = \begin{cases} 
1 - (\frac{x}{\xi})^2~\text{, if}~x<\xi,\\
0~\text{, otherwise}
\end{cases} \label{eq:rho}
\end{equation}

This formula indicates that two points located within a distance threshold \(\xi\) of each other are strongly correlated, whereas those beyond this threshold are considered independent. It is noteworthy that this term bears resemblance to the correlation function \(\rho(x) = \exp(-\frac{x}{\xi})\), which is commonly applied in statistical physics \cite{sethna2021statistical}, and $\xi$ represents the correlation length. 
Empirical evaluations indicate that the use of either correlation function expression significantly outperforms the scenario where all rays are assumed to be independent ($\rho=0$). Notably, a marginal improvement was observed when utilizing Eq.~\eqref{eq:rho}.

Therefore, we can approximate the correlation between two adjacent rays based on their expected depth, expressed as  $\rho_{mn} = \rho(d_{mn} \Delta \phi)$, where $\Delta \phi$ is the angular resolution of each pixel, $d_{mn}$ is the expected depth of ray at pixel $(m,n)$. This implies that when the camera is closer to an object, the observations in adjacent pixels of the camera exhibit stronger correlation. Hence the actual total information gain is smaller than the sum of the information gain of each pixel.
Accordingly, the correction function $f_{corr}$ in Eq.~\eqref{eq:id} can be defined as:
\begin{equation}
f_{corr}(\mathcal{H}(\bm{Z}_{mn}); d_{mn}) = \rho(d_{mn} \Delta \phi) \mathcal{H}(\bm{Z}_{mn})
\end{equation}
In our experiments, we let the correlation length $\xi=k D {\Delta \phi}$, where $D$ represents the diameter of the coarse bounding box enclosing the object, and $k$ is a hyperparameter, and we let $k=0.25$.

\textbf{Entropy of GMM}. We then introduce the details to compute the entropy for each ray, which is modeled as a Gaussian Mixture Model (GMM).
For the sake of simplicity, we denote the GMM's distribution as \(p(\bm{x}) = \sum_i w_i \mathcal{N}(\bm{x};\bm{\mu}_i, \bm{Q}_i)\).
We use the upper bound proposed in~\cite{huber2008entropy} to closely approximate the entropy of the GMM $\mathcal{H}(\bm{X})$:
\begin{equation}
\mathcal{H}(\bm{X}) \leq \sum_i w_i \left(- \log w_i + \frac{1}{2} \log \left( (2 \pi e)^N |\bm{Q}_i|\right) \right)~\label{eq:huber}
\end{equation}

This upper bound is expected to provide a more accurate approximation of the true entropy of the GMM compared to the conventional method which approximates the entropy using a single Gaussian that matches the first two moments of the GMM\cite{hershey2007approximating}, given by $\mathcal{H}(\bm{X}) \leq \frac{1}{2} \log \left( (2 \pi e)^N |\bm{\Sigma}|\right)$
where \(\bm{\Sigma}\) is calculated as:
\begin{equation}
\bm{\Sigma} = \sum_i w_i \left(\bm{Q}_i + (\bm{\mu_i}-\bm{\bar{\mu}})(\bm{\mu_i} - \bm{\bar{\mu}})^T \right)~\label{eq:single}
\end{equation}
and \(\bm{\bar{\mu}}\) is the weighted mean of the Gaussian components, defined as \(\bm{\bar{\mu}} = \sum_i w_i \bm{\mu_i} \).
It is worth mentioning that the baseline method 
\cite{neurar,pan2022activenerf} 
use the weighted average of position-based color variance to approximate the rays-based observation variance by
employing a single Gaussian whose mean and variance are the weighted averages of the means and variances of all samples along the rays, respectively ; in other words, $\bm{\Sigma} = \sum_i w_i \bm{Q}_i$. This approach resembles the first term in Eq.~\eqref{eq:single} but misses the covariance term \( (\bm{\mu_i}-\bm{\bar{\mu}})(\bm{\mu_i} - \bm{\bar{\mu}})^T \). Additionally, it does not take into account the visibility to the training views.

In summary, we derive an upper bound for the pixel-wise entropy, and consequently, for the joint entropy of each view, this upper bound is utilized to closely approximate the information gain at a given pose.
In the planning phase, given a candidate pose, we first apply Appendix Eq.~\eqref{eq:huber} to compute the entropy for each ray, subsequently, we compute the joint entropy of the image observation as per Eq.~\eqref{eq:id} at that pose, which then serves as a reward function in the planning process.

\subsection{Active mapping implementation details}

\textbf{Active Mapping Pipeline}. To train NVF within an active mapping framework, we build our pipeline on top of nerfstudio \cite{nerfstudio} and NerfBridge~\cite{yu2023nerfbridge}. 
Every time a new view is added to NVF, the model is trained from scratch on the collection of its observed views. 

After training, we sample candidate poses in the scene, without collision with the object, by filtering all poses within a density threshold.
In the Room scene, the sampler additionally thresholds for collisions between view poses and the current pose, to make sure the agent could move to the new pose without collision.
After candidate view poses are generated, NVF computes the entropy of each pose. The view with the highest entropy is next rendered in the scene and added to the observations. This procedure repeats until the horizon step is met, as is shown in
Alg.~\ref{algo:active_mapping}. 
In the experiments, we sample $N=512$ candidate views and run the active mapping for 20 steps; the evaluations are performed after the last planning step.

\textbf{Gradient-based Optimization for Planning}. In addition to the method of finding the best view among a randomly sampled candidate poses set, we also performed experiments on 6 DoF pose-refinement on the camera poses, $\bm{p} \in SE(3)$, as the entropy function $\mathcal{H}$ is a fully differentiable differentiable function of $\bm{p}$. We find the optimal $\bm{p}$ such that
\begin{equation}
    \bm{p}^* =\argmax_{\bm{p} \in SE(3)} \mathcal{H}(\bm{\bm{Z}_p})
\end{equation}
We first find the top $k$ poses with the highest entropy $\mathcal{P}_k$ and perform gradient-based optimization to refine the poses. To reduce the size of the computation graph and the memory requirements, a subset of pixels $\bm{Z}_i \subset \bm{Z_p}$ with an image is used to estimate the expected entropy, 
instead of the full image. We perform backpropogation on this estimated entropy using an Adam optimizer with a learning rate of $1e-4$, to find the optimum pose.

\begin{figure*}
    \centering
    \includegraphics[width=0.9\linewidth]{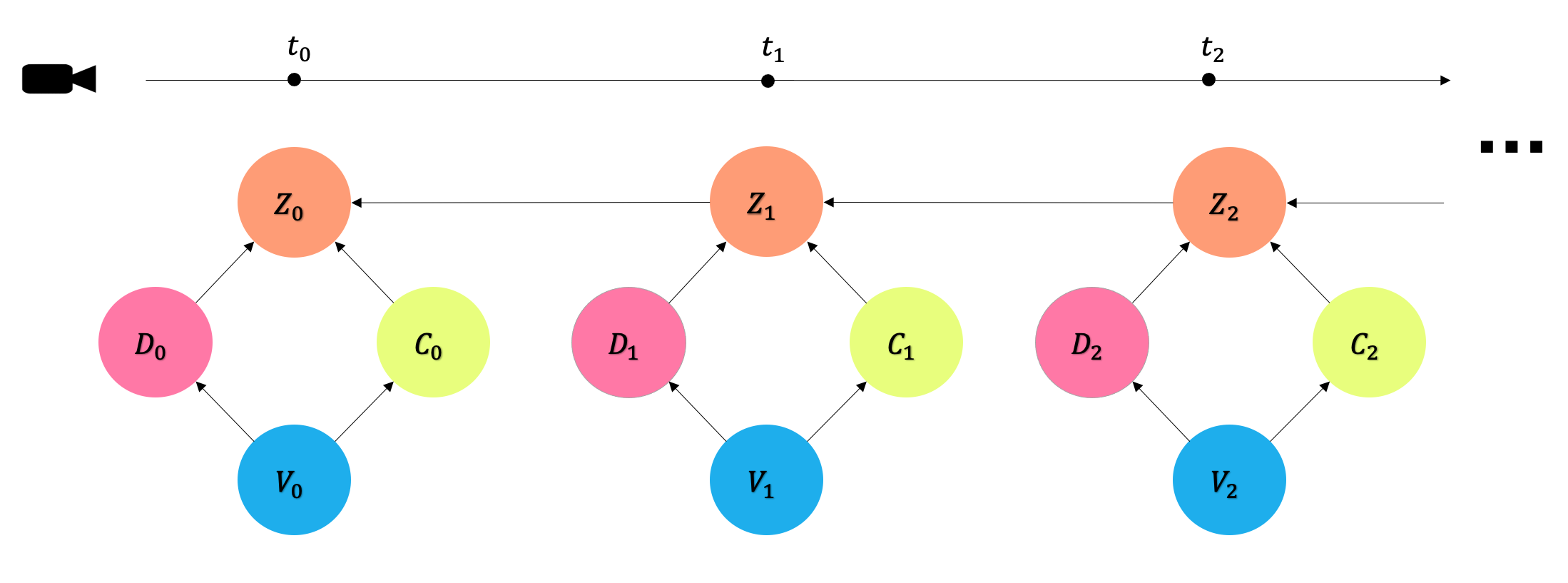}
    \caption{Schematics of the proposed Bayesian Network: $Z$ represents the observed (ray-based) color, $C$ represents the emitted (position-based) color, $D$ represents if the interval is occluded, $V$ represents the visibility, } 
    \label{fig:bayes}.
\end{figure*}

\begin{algorithm}
\caption{Active Mapping with NVF}\label{algo:active_mapping}
\begin{algorithmic}[1]

\State \textbf{Input:}

\State \; \; \; $ \mathcal{P} \gets \text{initial poses} $

\State \; \; \; $ \bm{Z} \gets \text{initial images}$

\For{$i=1$ to $n_{horizon}$}
\State $F_\Theta \gets$ trainNVF($\mathcal{P}, \bm{Z}$) \Comment{train NVF}

\State $\mathcal{P}_{c} \gets\text{samplePoses}(F_\Theta) $ \Comment{sample candidate poses} %

\State $\bm{p}_i \gets \argmax\limits_{\bm{p} \in \mathcal{P}_{c}}\mathcal{H} (\bm{Z_p}|F_\Theta ) $
\State $\mathcal{P} \gets \{\bm{p}_i\} \cup \mathcal{P}$
\State $\bm{Z} \gets \text{takeImageAt($\{\bm{p}_i\}$)} \cup \bm{Z}$ \Comment{update training set} %

\EndFor

\State \textbf{return} $F_\Theta$
\end{algorithmic}
\end{algorithm}

\begin{algorithm}
\caption{Gradient-Based Optimization for Planning}\label{algo:gradient_opt}
\begin{algorithmic}[1]
\State \textbf{Input:}
\State \;\;\;\;$ \mathcal{P} \gets \text{sampled poses}$
\State $\mathcal{P}_k \gets$ getTopKPoses$(\mathcal{P},\mathcal{H})$
\For{$i=1$ to $n_{iterations}$}
\For{$\bm{p}$ in $\mathcal{P}_k$}
\State $\bm{Z_p} \gets $sampleRays$(\bm{p})$ 
\State $\bm{p} \gets \bm{p} + \eta \frac{\partial (\mathcal{H}(\bm{Z_p}|F_\Theta))}{\partial \bm{p}}$
\EndFor
\EndFor
\State $\Tilde{\bm{p}} =  \argmax\limits_{\bm{p} \in \mathcal{P}_{k}}\mathcal{H} (\bm{Z_p}| F_\Theta ) $
\State \textbf{return} $\Tilde{\bm{p}}$
\end{algorithmic}
\end{algorithm}

\begin{figure*}
    \centering
    \includegraphics[width=0.9\textwidth]{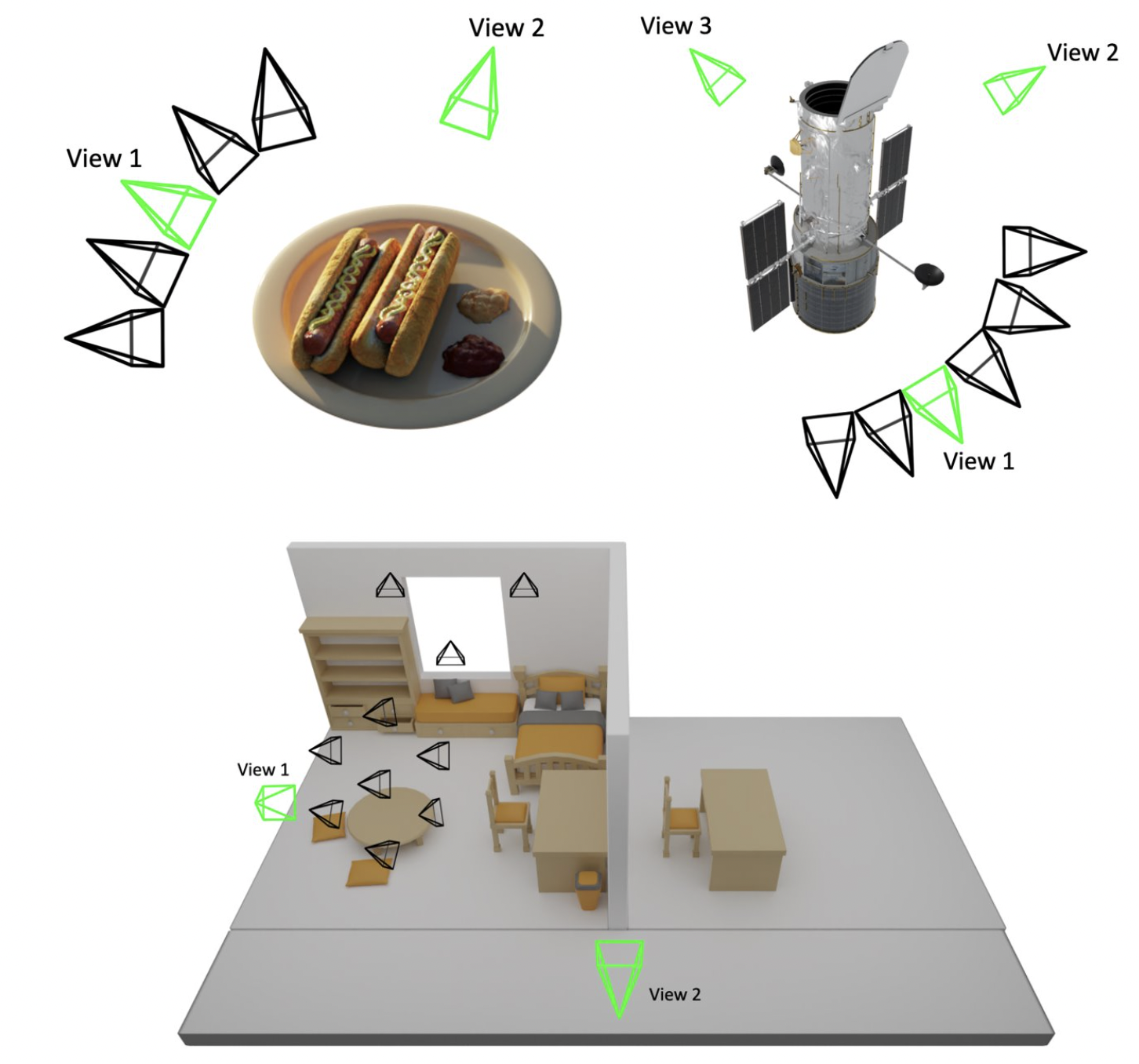}
    \caption{\textbf{Uncertainty Experiment Scene Setups:} 
Illustration of the training views and evaluation views in Fig.~\ref{fig:entropy_qualitative}. The black frustums correspond to the training views, the green frustums are the evaluation views.
For more video results, please refer to \url{https://sites.google.com/view/nvf-cvpr24/}
} 
    \label{fig:illustrative_fig}.
\end{figure*}

\section{Experiments Details}
\subsection{Uncertainty Estimation details}

As for the entropy comparison experiments shown in Fig.~\ref{fig:entropy_qualitative} of the main paper, Appendix Fig.~\ref{fig:illustrative_fig} provides an illustration of the pose of the training views and evaluation views.

\subsection{Mesh metrics implementation details}
For computing Accuracy, Completion, and Completion Ratio metrics, ground truth points are sampled from the ground truth scene meshes. Points from NVF's reconstructed mesh are sampled from the observation view rays. Accuracy measures the mean distance of sampled points from the reconstructed mesh to the nearest corresponding points in the ground truth mesh. Completion instead measures the mean distance of sampled ground truth points to the nearest reconstructed mesh points. Completion Ratio calculates the percentage of completion distances being below a threshold. For the original NeRF assets and Hubble scene, the threshold is set to 0.01. For the Room scene, as the scale is larger, the threshold is set to 0.1.

Visual coverage quantifies the surface area a trajectory of views covers a scene. We compute this with rasterization. Given a ground truth mesh of the scene, we project the mesh onto all of the observation views. In each rendered image, we record the number of mesh faces visible to the corresponding view. We append all observed faces to a visible set. Computing visual coverage is then the ratio of the length of the visible set to the total number of faces in the mesh.

\section{More Qualitative Results}

\begin{figure*}
    \centering
    \includegraphics[width=0.8\textwidth]{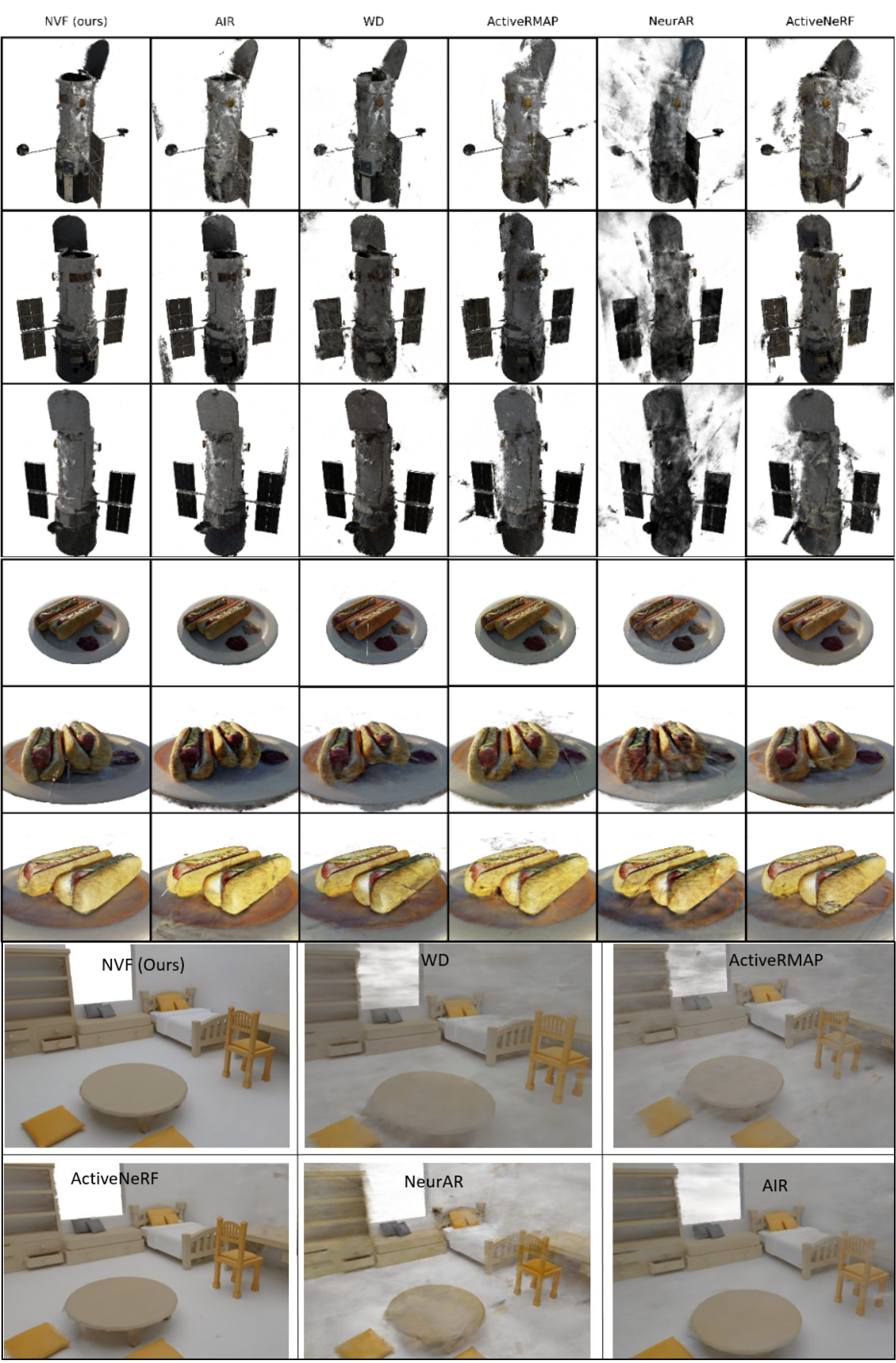}
    \caption{\textbf{Qualitative Results:} Comparisons on novel view synthesis results. Our method demonstrates superior novel-view synthesis rendering fine details in comparison to all baselines. For more video results, please refer to \url{https://sites.google.com/view/nvf-cvpr24/}} 
    \label{figa:qual_fig}
\end{figure*}
\subsection{Active Mapping}

In addition to the results in Tab.~\ref{table:1}, more qualitative results are presented in Appendix Fig.~\ref{figa:qual_fig}. As shown, our method achieves better novel view synthesis quality compared to baseline methods.

\begin{figure*}
    \centering
    \includegraphics[width=0.9\textwidth]{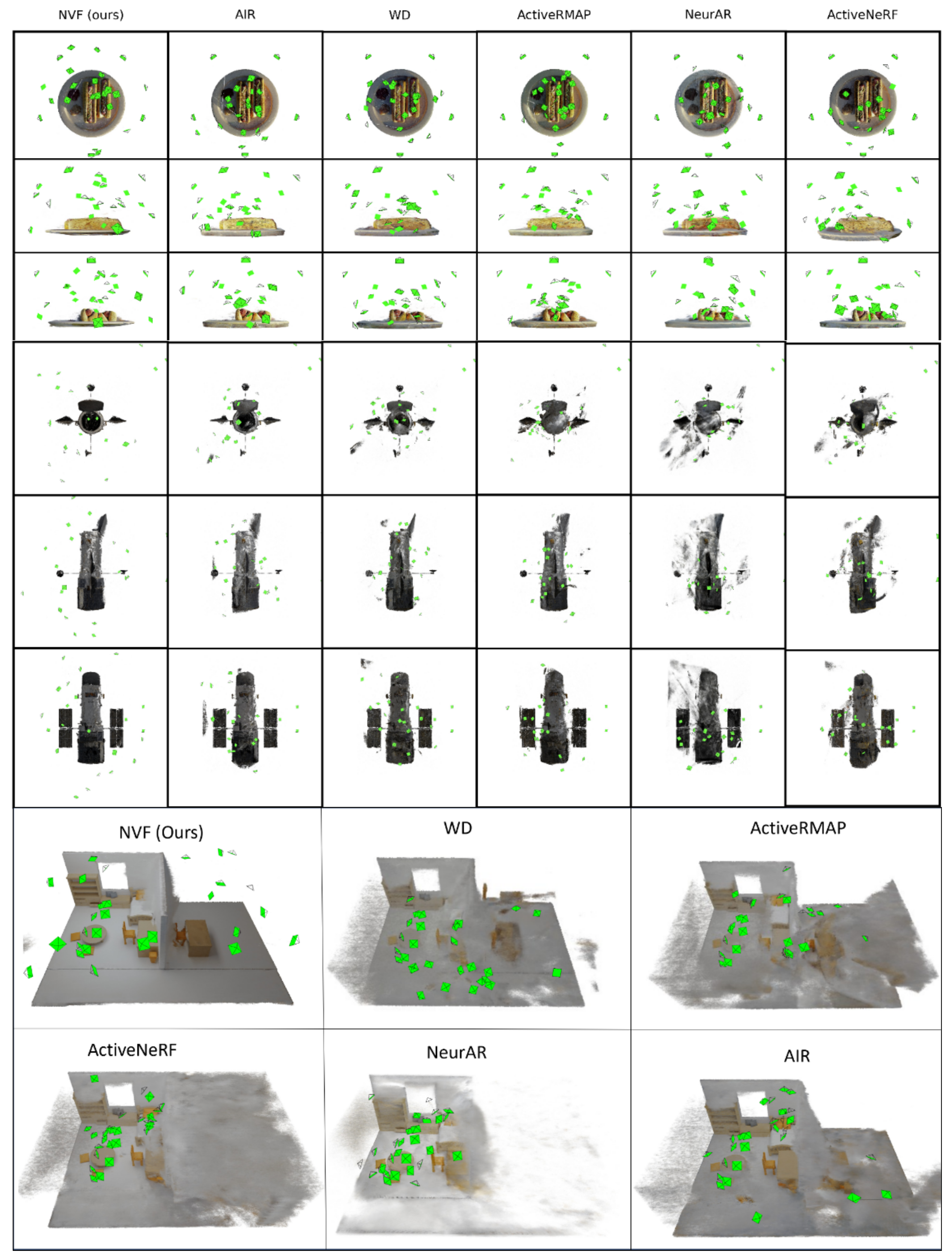}
    \caption{\textbf{Additional reconstruction results and camera view distribution} For more video results, please refer to \url{https://sites.google.com/view/nvf-cvpr24/} } 
    \label{fig:experiment_schematic}.
\end{figure*}

\begin{table*}[htbp]
  \centering
  \caption{Performance of gradient-based methods} 
    \begin{tabular}{ccccccccc}
    \toprule
    Method & PSNR$\uparrow$ & SSIM$\uparrow$ & LPIPS$\downarrow$ & RGB$\downarrow$ & Acc.$\downarrow$ & Comp.$\downarrow$ & C.R.$\uparrow$ & Vis.$\uparrow$ \\
    \midrule
    AIR   & 24.63 & 0.862 & 0.182 & 0.0035 & 0.0249 & 0.0140 & 0.525 & 0.586 \\
    NeurAR & 25.19 & 0.772 & 0.265 & 0.0030 & 0.0480 & 0.0170 & 0.416 & 0.537 \\
    NVF   & 27.99 & 0.919 & 0.100 & 0.0016 & 0.0225 & 0.0110 & 0.651 & 0.681 \\
    \midrule
    AIR-OPT & 24.41 & 0.858 & 0.183 & 0.0037 & 0.0267 & 0.0159 & 0.450 & 0.548 \\
    NeurAR-OPT & 25.42 & 0.794 & 0.245 & 0.0029 & 0.0461 & 0.0180 & 0.381 & 0.563 \\
    NVF-OPT  & \textbf{29.33} & \textbf{0.930} & \textbf{0.086} & \textbf{0.0012} & \textbf{0.0196} & \textbf{0.0106} & \textbf{0.666} & \textbf{0.690} \\
    \bottomrule
    \end{tabular}%
  \label{tab:grad}%
\end{table*}%

\subsection{Gradient-based Pose-Optimization results}

Certain methods compare uncertainty among a finite set of pre-defined scene-specific view candidates. This limits their applicability to previously unseen scenes as well as their ability to reach an optimal solution. Gradient-based pose estimation aims to find the next-best-view (NBV) on a continuous manifold which broadens its applicability to different scenarios and results in optimal view selection.

The results in Tab.~\ref{table:1} highlight our approach's ability to select the optimal view from proposed candidates, intentionally omitting gradient-based optimization to ensure a fair comparison. To extend our analysis, we conducted a further comparison with gradient-based optimization methods for view selection, detailed in Tab.~\ref{tab:grad}. This comparison, which includes our method and two others~\cite{neurar, yan2023active}, utilizes gradient descent to refine the selection of views. As demonstrated in Appendix Tab.~\ref{tab:grad}, the integration of gradient-based optimization considerably improves our method's performance, allowing it to surpass competing gradient-based approaches. This superior performance is attributed to our method's more precise estimation of uncertainty.

\subsection{Additional Results}
We present the complete results of all original NeRF assets in Appendix Tab.~\ref{table:nerf_asset_result1} \&~\ref{table:nerf_asset_result2}.
We also present the complete results of the ablation study in Appendix Tab.~\ref{table:ablation_supp}. The result is averaged across all scenes.

\begin{table*}[htbp]
  \centering
  \caption{Ablation Studies} \label{table:ablation_supp}
        \begin{tabular}{ccccccccc}
    \toprule
    Ablations & PSNR$\uparrow$ & SSIM$\uparrow$ & LPIPS$\downarrow$ & RGB$\downarrow$ & Acc.$\downarrow$ & Comp.$\downarrow$ & C.R.$\uparrow$ & Vis.$\uparrow$ \\
    \midrule
    w/o Vis. & 21.11 & 0.844 & 0.187 & 0.0119 & 0.0466 & 0.0765 & 0.479 & 0.382 \\
w/o Var. & 23.77 & 0.897 & 0.113 & 0.0049 & 0.0276 & 0.0305 & 0.639 & 0.551 \\
Ind. Rays & 20.32 & 0.822 & 0.236 & 0.0125 & 0.0560 & 0.0506 & 0.451 & 0.482 \\
Loose & 22.54 & 0.881 & 0.137 & 0.0100 & 0.0247 & 0.0609 & 0.600 & 0.504 \\
NVF (Ours) & 24.42 & 0.902 & 0.108 & 0.0041 & 0.0287 & 0.0324 & 0.628 & 0.546 \\
    \bottomrule
    \end{tabular}%
  \label{tab:ablation_result_supp}%
\end{table*}%

\begin{table*}[htbp]
  \centering
  \caption{Results of original NeRF assets (1)} \label{table:nerf_asset_result1}
\begin{tabular}{cccccccccc}
\hline
Scene & Method & PSNR$\uparrow$ & SSIM$\uparrow$ & LPIPS$\downarrow$ & RGB$\downarrow$ & Acc.$\downarrow$ & Comp.$\downarrow$ & C.R.$\uparrow$ & Vis.$\uparrow$ \\
\hline
\multirow{7}[2]{*}{Chair} & Random & 17.17 & 0.835 & 0.190 & 0.0193 & 0.0470 & 0.0470 & 0.250 & 0.311 \\
      & WD    & 18.07 & 0.853 & 0.197 & 0.0163 & 0.0386 & 0.0167 & 0.499 & 0.582 \\
      & ActiveRMAP & 18.67 & 0.863 & 0.183 & 0.0136 & 0.0277 & 0.0144 & 0.584 & 0.614 \\
      & AIR   & 18.47 & 0.859 & 0.176 & 0.0155 & 0.0296 & 0.0135 & 0.568 & 0.614 \\
      & ActiveNeRF & 15.90 & 0.806 & 0.257 & 0.0280 & 0.0295 & 0.0223 & 0.407 & 0.503 \\
      & NeurAR & 19.24 & 0.817 & 0.231 & 0.0127 & 0.0427 & 0.0155 & 0.485 & 0.596 \\
      & NVF (Ours) & \textbf{23.89} & \textbf{0.937} & \textbf{0.057} & \textbf{0.0041} & \textbf{0.0209} & \textbf{0.0089} & \textbf{0.763} & \textbf{0.705}\\
\hline
\multirow{7}[2]{*}{Drums} & Random & 17.08 & 0.753 & 0.286 & 0.0198 & 0.0378 & 0.0162 & 0.518 & 0.193 \\
      & WD    & 19.07 & 0.796 & 0.252 & 0.0126 & 0.0288 & 0.0130 & 0.575 & 0.444 \\
      & ActiveRMAP & 18.77 & 0.784 & 0.264 & 0.0134 & 0.0385 & 0.0128 & 0.574 & 0.443 \\
      & AIR   & 19.00 & 0.789 & 0.277 & 0.0126 & 0.0319 & 0.0115 & 0.596 & 0.464 \\
      & ActiveNeRF & 18.35 & 0.767 & 0.305 & 0.0147 & 0.0325 & 0.0160 & 0.479 & 0.393 \\
      & NeurAR & 18.22 & 0.722 & 0.328 & 0.0151 & 0.0434 & 0.0158 & 0.453 & 0.401 \\
      & NVF (Ours) & \textbf{21.00} & \textbf{0.866} & \textbf{0.142} & \textbf{0.0079} & \textbf{0.0186} & \textbf{0.0069} & \textbf{0.836} & \textbf{0.541} \\
\hline
\multirow{7}[2]{*}{Ficus} & Random & 19.86 & 0.826 & 0.202 & 0.0103 & 0.0254 & 0.0141 & 0.671 & 0.355 \\
      & WD    & 17.98 & 0.777 & 0.316 & 0.0163 & 0.0299 & 0.0172 & 0.553 & 0.601 \\
      & ActiveRMAP & 19.40 & 0.803 & 0.263 & 0.0122 & 0.0260 & 0.0122 & 0.653 & 0.637 \\
      & AIR   & 18.75 & 0.772 & 0.325 & 0.0134 & 0.0237 & 0.0145 & 0.575 & 0.554 \\
      & ActiveNeRF & 18.75 & 0.762 & 0.366 & 0.0134 & 0.0210 & 0.0202 & 0.560 & 0.529 \\
      & NeurAR & 20.27 & 0.755 & 0.337 & 0.0094 & 0.0254 & 0.0189 & 0.545 & 0.513 \\
      & NVF (Ours) & \textbf{22.76} & \textbf{0.900} & \textbf{0.089} & \textbf{0.0053} & \textbf{0.0112} & \textbf{0.0062} & \textbf{0.896} & \textbf{0.649} \\
\hline
\multirow{7}[2]{*}{Hotdog} & Random & 19.87 & 0.861 & 0.166 & 0.0107 & 0.0379 & 0.0565 & 0.239 & 0.361 \\
      & WD    & 21.84 & 0.892 & 0.131 & 0.0066 & 0.0186 & 0.0395 & 0.344 & 0.455 \\
      & ActiveRMAP & 22.75 & 0.895 & 0.130 & 0.0053 & 0.0197 & 0.0415 & 0.338 & 0.466 \\
      & AIR   & 22.35 & 0.897 & 0.124 & 0.0058 & 0.0197 & 0.0381 & 0.351 & 0.470 \\
      & ActiveNeRF & 21.57 & 0.885 & 0.145 & 0.0070 & 0.0234 & 0.0335 & 0.324 & 0.461 \\
      & NeurAR & 22.90 & 0.866 & 0.171 & 0.0051 & 0.0279 & \textbf{0.0320} & 0.317 & 0.450 \\
      & NVF (Ours) & \textbf{26.10} & \textbf{0.928} & \textbf{0.084} & \textbf{0.0025} & \textbf{0.0157} & 0.0356 & \textbf{0.371} & \textbf{0.472} \\
\hline
\end{tabular}%
\end{table*}%

\begin{table*}[htbp]
  \centering
  \caption{Results of original NeRF assets (2)} \label{table:nerf_asset_result2}
\begin{tabular}{cccccccccc}
\hline
Scene & Method & PSNR$\uparrow$ & SSIM$\uparrow$ & LPIPS$\downarrow$ & RGB$\downarrow$ & Acc.$\downarrow$ & Comp.$\downarrow$ & C.R.$\uparrow$ & Vis.$\uparrow$ \\
\hline
\multirow{7}[2]{*}{Lego} & Random & 16.49 & 0.720 & 0.265 & 0.0229 & 0.0599 & 0.0504 & 0.161 & 0.115 \\
      & WD    & 18.54 & 0.771 & 0.217 & 0.0142 & 0.0305 & 0.0283 & 0.257 & 0.224 \\
      & ActiveRMAP & 17.49 & 0.752 & 0.234 & 0.0180 & 0.0238 & 0.0237 & 0.280 & 0.227 \\
      & AIR   & 19.33 & 0.797 & 0.189 & 0.0118 & 0.0262 & 0.0249 & 0.296 & 0.230 \\
      & ActiveNeRF & 17.59 & 0.736 & 0.263 & 0.0176 & 0.0265 & 0.0317 & 0.222 & 0.199 \\
      & NeurAR & 15.12 & 0.713 & 0.277 & 0.0314 & 0.0246 & 0.0357 & 0.319 & 0.189 \\
      & NVF (Ours) & \textbf{23.97} & \textbf{0.896} & \textbf{0.082} & \textbf{0.0040} & \textbf{0.0131} & \textbf{0.0167} & \textbf{0.426} & \textbf{0.270} \\
\hline
\multirow{7}[2]{*}{Materials} & Random & 15.90 & 0.802 & 0.220 & 0.0266 & 0.0409 & 0.0800 & 0.117 & 0.089 \\
      & WD    & 19.38 & 0.845 & 0.174 & 0.0122 & 0.0197 & 0.0275 & 0.343 & 0.304 \\
      & ActiveRMAP & 19.68 & 0.843 & 0.174 & 0.0117 & 0.0213 & 0.0271 & 0.345 & 0.303 \\
      & AIR   & 19.45 & 0.844 & 0.171 & 0.0138 & 0.0238 & 0.0320 & 0.318 & 0.289 \\
      & ActiveNeRF & 18.73 & 0.833 & 0.191 & 0.0135 & 0.0207 & 0.0290 & 0.322 & 0.287 \\
      & NeurAR & 19.68 & 0.833 & 0.182 & 0.0109 & 0.0196 & 0.0339 & 0.348 & 0.255 \\
      & NVF (Ours) & \textbf{25.36} & \textbf{0.931} & \textbf{0.061} & \textbf{0.0029} & \textbf{0.0107} & \textbf{0.0134} & \textbf{0.564} & \textbf{0.396} \\
\hline
\multirow{7}[2]{*}{Mic} & Random & 21.18 & 0.851 & 0.205 & 0.0081 & 0.0294 & 0.0276 & 0.468 & 0.257 \\
      & WD    & 26.79 & 0.942 & 0.067 & 0.0022 & 0.0176 & 0.0087 & 0.755 & 0.564 \\
      & ActiveRMAP & 26.60 & 0.940 & 0.069 & 0.0022 & 0.0187 & 0.0095 & 0.752 & 0.532 \\
      & AIR   & 24.81 & 0.927 & 0.107 & 0.0034 & 0.0165 & 0.0091 & 0.728 & 0.508 \\
      & ActiveNeRF & 24.96 & 0.926 & 0.101 & 0.0033 & 0.0198 & 0.0105 & 0.709 & 0.497 \\
      & NeurAR & 25.15 & 0.889 & 0.159 & 0.0031 & 0.0304 & 0.0099 & 0.679 & 0.528 \\
      & NVF (Ours) & \textbf{27.99} & \textbf{0.956} & \textbf{0.053} & \textbf{0.0016} & \textbf{0.0161} & \textbf{0.0070} & \textbf{0.854} & \textbf{0.566} \\
\hline
\multirow{7}[2]{*}{Ship} & Random & 15.75 & 0.578 & 0.483 & 0.0281 & 0.0580 & 0.0456 & 0.250 & 0.252 \\
      & WD    & 19.54 & 0.663 & 0.369 & 0.0112 & 0.0525 & 0.0313 & 0.374 & 0.540 \\
      & ActiveRMAP & 19.61 & 0.665 & 0.343 & 0.0112 & 0.0487 & 0.0290 & 0.385 & 0.543 \\
      & AIR   & 19.22 & 0.658 & 0.367 & 0.0121 & 0.0515 & 0.0307 & 0.378 & 0.513 \\
      & ActiveNeRF & 17.19 & 0.569 & 0.485 & 0.0197 & 0.0606 & 0.0370 & 0.329 & 0.496 \\
      & NeurAR & 19.38 & 0.556 & 0.491 & 0.0115 & 0.0569 & 0.0461 & 0.331 & 0.483 \\
      & NVF (Ours) & \textbf{22.32} & \textbf{0.742} & \textbf{0.254} & \textbf{0.0059} & \textbf{0.0445} & \textbf{0.0188} & \textbf{0.454} & \textbf{0.596} \\
\hline
\end{tabular}%
\end{table*}%
 \fi
\ifreview \clearpage \appendix  \fi

\end{document}